\documentclass[10pt,a4paper,twocolumn,english]{article}
\usepackage{sbsr}

\usepackage{multirow} 
\usepackage{array} 
\usepackage{makecell} 
\usepackage{paralist} 

\title{Mapping Rio de Janeiro's Favelas: General-Purpose vs. Satellite-Specific Neural Networks}

\author{Thomas Hallopeau$^1$, Joris Guérin$^1$, Laurent Demagistri$^{1, 4}$, Youssef Fouzai$^1$, Renata Gracie$^{2, 4}$, Vanderlei Pascoal De Matos$^{2, 4}$, Helen Gurgel$^{3, 4}$, Nadine Dessay$^{1, 4}$}

\address{$^1$Espace-Dev, French National Research Institute for Sustainable Development, University of Montpellier, France, firstname.lastname@ird.fr, $^2$Institute of Scientific and Technological Communication and Information in Health, Fiocruz, Rio de Janeiro, Brazil, $^3$Department of Geography, University of Brasilia, Brazil, $^4$International Joint Laboratory "Sentinela"}

\renewcommand{\bfseries}{\scshape}

\renewcommand{\bfseries}{\normalfont}

\usepackage[scaled=0.92]{helvet} 

\usepackage{booktabs}

\begin{document}

\twocolumn[
\begin{@twocolumnfalse}
\maketitle
\end{@twocolumnfalse}
]


\section*{Abstract} 

\hspace{-1.5mm}\textit{ 
While deep learning methods for detecting informal settlements have already been developed, they have not yet fully utilized the potential offered by recent pretrained neural networks. We compare two types of pretrained neural networks for detecting the favelas of Rio de Janeiro: \newline 1. Generic networks pretrained on large diverse datasets of unspecific images, 2. A specialized network pretrained on satellite imagery. While the latter is more specific to the target task, the former has been pretrained on significantly more images. Hence, this research investigates whether task specificity or data volume yields superior performance in urban informal settlement detection.
}
\\

\textit{\textbf{Key words --} 
remote sensing foundation models, intra-urban classification, informal settlements. 
}

\section{Introduction}




Spontaneous settlements develop rapidly and outside of a formal legal framework, making them difficult to map with field surveys or socio-economic data. Automatic mapping of informal settlements initially relied on object-oriented image analysis approaches, based on ontologies, using very high-resolution imagery~\cite{hofmannDetectingInformalSettlements2008,kohliOntologySlumsImagebased2012}. Later, machine learning approaches emerged, based on texture and morphology or integrating various spectral indices~\cite{kufferExtractionSlumAreas2016,wurmExploitationTexturalMorphological2017}. However, selecting and computing the variables required for these approaches is both complex and time-consuming.  Deep learning addresses this challenge by leveraging the ability of neural networks to automatically learn and extract relevant features from raw data, particularly from images~\cite{alzubaidiReviewDeepLearning2021}. Many deep learning methods have been developed for detecting informal settlements. Most of these approaches train convolutional or fully connected neural networks from scratch to segment and detect informal settlements in medium- or very high-resolution satellite images~\cite{perselloDeepFullyConvolutional2017, hafnerUnsupervisedDomainAdaptation2022}. In 2019, Stark et al.~\cite{starkSlumMappingImbalanced2019} employed a fully connected network pretrained on natural images to detect informal settlements across various imbalanced datasets. Since then, no studies have fully explored the potential of: 

\begin{itemize}
    \item Latest pretrained CNNs (Convolutional Neural Networks) and ViTs (Vision Transformers)
    \item Remote Sensing Foundation Models (RSFM), neural networks pretrained specifically on satellite imagery
\end{itemize}

In this paper, we propose to compare the features extracted by pretrained CNNs and ViTs on unspecialized and large image databases from the internet such as ImageNet~\cite{dengImageNetLargescaleHierarchical2009} with those extracted by CROMA~\cite{fullerCROMARemoteSensing2023}, a specialized network pretrained on a more limited set of satellite imagery. This paper represents the first attempt at utilizing RSFM for informal settlement mapping. The pretrained models are compared at the task of classifying tiles into binary categories: favela vs. non-favela.


\section{Methods}


\subsection{Data description}
\label{subsec:RG}

\begin{figure}[t] 
    \begin{center}
    \includegraphics[width=\linewidth]{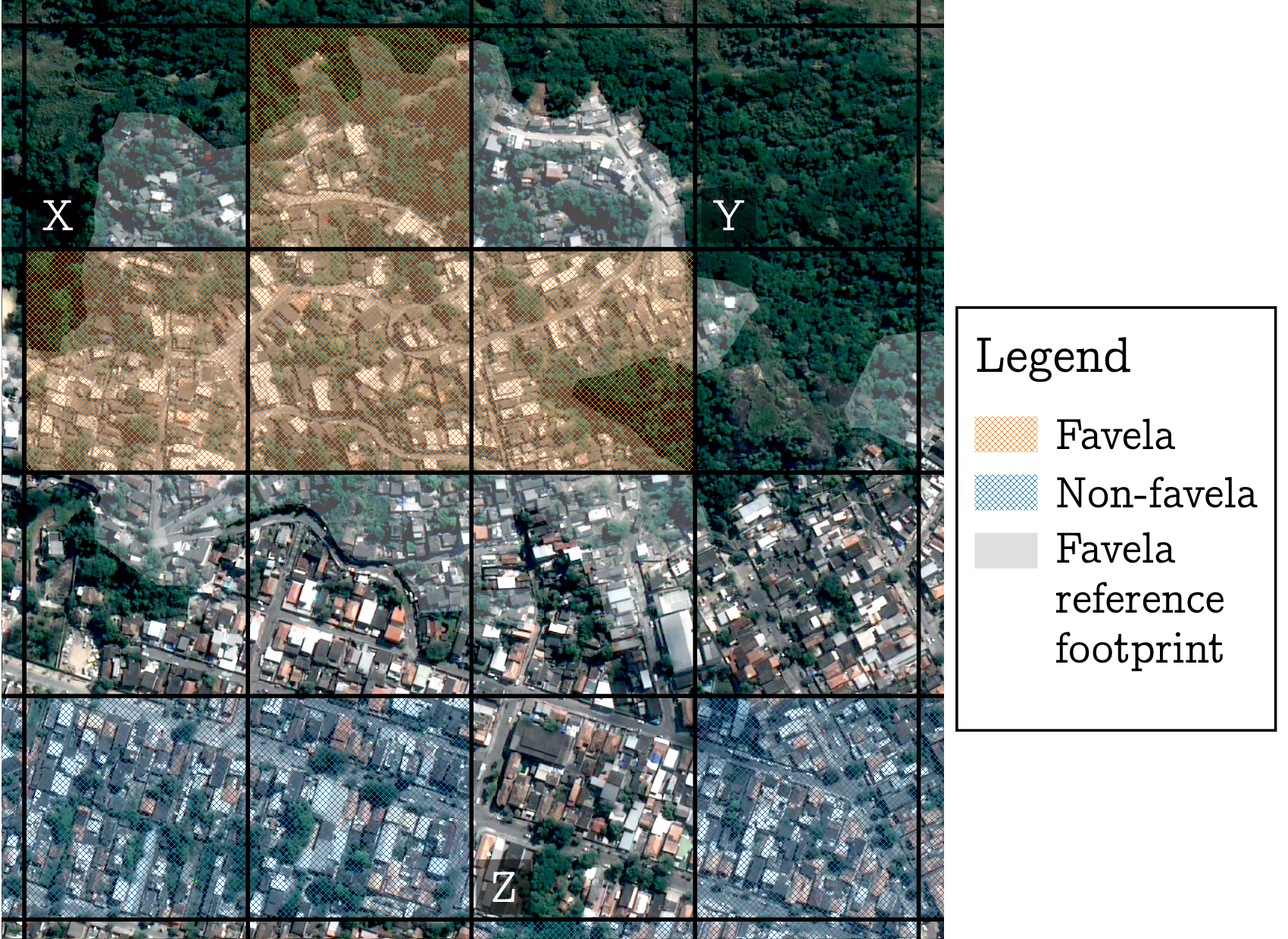}
    \caption{Favela and non-favela tiles. Non-retained tiles are not shaded.}
    \label{fig:mesh}
    \end{center}
\end{figure}

The study area corresponds to the administrative boundaries of the city of Rio de Janeiro and it was divided into tiles according to an orthogonal grid with a 150m resolution. For each tile, the following variables were calculated in order to create a labeled dataset:
\begin{itemize}
    \item The proportion of surface covered by favelas: computed from the favela outlines provided as a multi-polygon by the Brazilian Institute of Geography and Statistics (IBGE), 
    \item The proportion of surface covered by vegetation: based on the Normalized Difference Vegetation Index (NDVI) 
    computed from a Pléiades image and thresholded at 0.6 (each pixel with an NDVI greater than or equal to 0.6 is associated 
    with vegetation; this threshold was adjusted through photo-interpretation),
    \item The proportion of surface covered by buildings: based on data from the Global Human Settlement Layer (GHSL) of the European Commission.
\end{itemize}
Based on OpenStreetMap data, the presence or absence of an industrial zone within each tile was also recorded. Then, tiles meeting at least one of the following conditions were removed from the dataset:
\begin{itemize}
    \item Proportion of surface covered by buildings strictly less than 50\%,
    \item Proportion of surface covered by vegetation strictly greater than 95\%,
    \item Presence of an industrial zone within the tile.
\end{itemize}
For example, tile Y in Figure~\ref{fig:mesh} was not retained because it is only composed of vegetation. tile Z was also not retained because it is associated with the presence of an industrial zone in OpenStreetMap. A class was assigned to some of the remaining tiles based on the following criteria:
\begin{itemize}
    \item \underline{Favela tiles} corresponding to favelas: the proportion of surface covered by favelas is greater than or equal to 70\%.
    \item \underline{Non-favela tiles} corresponding to non-favela urban areas, most often formal residential zones: the proportion of surface covered by favelas is zero.
\end{itemize}
Some tiles meet the initial selection criteria but were not associated with any class and were therefore also eliminated, see tile X in Figure~\ref{fig:mesh}. For this tile, the proportion of surface covered by favelas falls within $\left] 0, 70\% \right[$. This tile was not considered as non-favela because it is covered by a favela, but it is not sufficiently covered by the reference favela vector to be considered as favela.

\subsection{Neural networks selected for comparison}

In the PyTorch environment, it is possible to access a wide range (over a thousand) of generic pretrained neural networks using the PyTorch Image Models (timm) library~\cite{rw2019timm}. These neural networks are pretrained on extensive and non-specific image databases from the internet. We selected 2 neural networks for testing:
\begin{itemize}
    \item A ViT pretrained using self-supervised learning on LVD-142M\footnote{vit\_base\_patch14\_dinov2.lvd142m}~\cite{oquabDINOv2LearningRobust2023}, 
    \item A CNN pretrained using supervised learning on ImageNet-22k\footnote{convnext\_base.fb\_in22k}~\cite{dengImageNetLargescaleHierarchical2009}.
\end{itemize}
We used these neural networks with the RGB channels of a Pléiades image, leaving aside the infrared band.

Numerous RSFMs have been published since late 2021~\cite{wangAdvancingPlainVision2022,chaBillionscaleFoundationModel2023,guoSkySenseMultiModalRemote2023,hongSpectralGPTSpectralRemote2023}. These neural networks are often pretrained on fewer images compared to recent generic pretrained models but are specialized for aerial views. Some of them exploit the information available across all spectral bands of satellite images rather than relying solely on the RGB channels of traditional images. The most recent RSFMs typically feature ViT architectures and are trained in a self-supervised manner, especially through masked autoencoding~\cite{heMaskedAutoencodersAre2021}. They mainly differ in their self-supervised learning techniques and their pretraining datasets: ground resolution, spectral composition, dataset size, and geographic coverage. Among these neural networks, CROMA~\cite{fullerCROMARemoteSensing2023} stands out for two main reasons: 
\begin{itemize} 
    \item It operates with Sentinel-1 GRD and/or Sentinel-2 L2A images. The free availability of these images through the Copernicus portal\footnote{\href{https://browser.dataspace.copernicus.eu}{https://browser.dataspace.copernicus.eu}} can be beneficial for later establishing dynamic monitoring of favelas. 
    \item It is relatively easy to implement with PyTorch, using similar types of code as those required for generic pretrained neural networks from the timm library. 
\end{itemize} 
CROMA is pretrained on the SSL4EO dataset~\cite{wangSSL4EOS12LargeScaleMultiModal2022}, an unannotated dataset consisting of over one million pairs (Sentinel-1 GRD, Sentinel-2 L2A) of 264x264 pixel images from around the world. CROMA's self-supervised pretraining includes two components: 
\begin{itemize}
    \item Reconstruction learning: 75\% of each Sentinel-1 and Sentinel-2 image is masked, and the network must reconstruct the images
    \item Contrastive learning: The neural network learns to match pairs of Sentinel-1 and Sentinel-2 images (both images correspond to the same location and roughly the same time). 
\end{itemize}
CROMA's architecture consists of a ViT that encodes multispectral images, a ViT that encodes radar images, and a shared encoder that processes the outputs from both ViTs when working with both multispectral and radar images simultaneously.

\subsection{Feature extraction}

We tested the neural networks using feature extraction. The pretrained neural networks were employed to extract features from the tiles associated with the grid described in section~\ref{subsec:RG}. These raw outputs represent the tiles as vectors of values, encapsulating abstract and, in principle, relevant information about the tiles. We then used these outputs to train and test a random forest classifier~\cite{breimanRandomForests2001a}. This model performs well on heterogeneous data and helps mitigate the risk of overfitting, which is particularly suitable given the complex urban sprawl of Rio de Janeiro and the diverse characteristics of its favelas. Additionally, random forests are non-linear models, making them ideal for our intra-urban classification task, where some favelas may exhibit characteristics very similar to those of formal neighborhoods~\cite{fernandez-delgadoWeNeedHundreds2014a}.

\subsection{Data balancing}
\label{subsec:DB}

Balancing datasets is essential to prevent the error calculation in machine learning models from being skewed towards the majority class. Non-favela tiles were over-represented compared to favela tiles. Based on the previously mentioned filters (see section~\ref{subsec:RG}), there were approximately 30 times more non-favela tiles than class favela tiles in the global dataset. To mitigate this imbalance, the dataset was balanced through random undersampling~\cite{heLearningImbalancedData2009}: for each training or test set, an equal number of non-favela and favela tiles were randomly selected.

\subsection{Evaluation protocol}

We selected the following evaluation metrics to assess the detection quality achieved by our detection approach: Recall, Precision, and F1-score. Precision measures the model's ability to avoid making errors when predicting positive samples, while Recall measures the model's ability to correctly predict all positive samples. The F1-score reflects the overall quality of the prediction. These three evaluation metrics, commonly used in binary classification, provide a comprehensive view of detection quality.

Since favelas exhibit variable characteristics within the city of Rio de Janeiro, we evaluated our approach using 5-folds cross-validation. In each cross-validation, the data was randomly divided into 5 folds that were then balanced by random undersampling, as described in section~\ref{subsec:DB}. Successively, 4 subsets were used for training, and testing was performed on the fifth subset. This process was repeated 5 times so that each subset was used for testing once. An untrained model was used at each iteration.

In order to account for the randomness in the training process due to data balancing, we performed 5 successive 5-folds cross-validations for each network. This resulted in 25 values for each precision metric. We then computed the average metric values and the standard deviation for each metric.

\section{Results and discussion}


Our results are reported in Table~\ref{tab:results}. The results obtained from both general-purpose and satellite-specific pretrained neural networks demonstrate their potential for detecting informal settlements. The RSFM, CROMA, outperforms the generic pretrained neural networks. This underscores the value of domain-specific pretraining and highlights the effectiveness of satellite-specific models in detecting complex urban features, such as favelas in Rio de Janeiro, compared to generic models, despite the latter being pretrained on significantly more diverse datasets. Although generic pretrained networks have been applied to a Pléiades image, which offers a higher ground resolution compared to Sentinel images (5m versus 10m for the highest-resolution Sentinel bands), their reliance on only the three RGB channels and their lack of specificity appear to offset the advantages of their large pretraining datasets.


\begin{table}[h]
\centering
\begin{tabular}{lccc}
\toprule
\small Method & \small Precision & \small Recall & \small F1-score \\
\midrule
\small CROMA RSFM & $0.81 \pm 0.03$ & $0.81 \pm 0.05$ & $0.81 \pm 0.02$ \\
\small Generic ViT & $0.71 \pm 0.03$ & $0.75 \pm 0.06$ & $0.72 \pm 0.03$ \\
\small Generic CNN & $0.71 \pm 0.05$ & $0.74 \pm 0.09$ & $0.72 \pm 0.03$ \\
\bottomrule
\end{tabular}
\caption{Results obtained (mean and standard deviation) over 25 values from 5 successive 5-fold cross-validations for CROMA and the 2 generic pretrained neural networks.}
\label{tab:results}
\end{table}



	
\section{Conclusions}

We detected favelas by classifying tiles while comparing two types of pretrained neural networks: a Remote Sensing Foundation Model specifically pretrained on satellite images, and 2 generic models pretrained on large and significantly more diverse datasets of non-specific and general-purpose images. In both cases, the encouraging results demonstrate the potential of pretrained deep neural networks for detecting informal settlements,  presenting lower development costs compared with traditional methods. Furthermore, the fact that they are pretrained on diverse natural images or on satellite images of the entire Earth suggests their applicability to other informal settlements beyond those in Rio de Janeiro. Finally, the results emphasize the advantage of domain-specific pretraining with satellite imagery, with the Remote Sensing Foundation Model, CROMA, showing better results than its generic counterparts.



\renewcommand{\refname}{References}
{\small
\bibliographystyle{unsrt} 
\bibliography{07_SBSR.bib}}

\begin{thebibliography}{10}

\bibitem{hofmannDetectingInformalSettlements2008}
P.~Hofmann, J.~Strobl, T.~Blaschke, and H.~Kux.
\newblock Detecting informal settlements from {{QuickBird}} data in {{Rio}} de {{Janeiro}} using an object based approach.
\newblock In Thomas Blaschke, Stefan Lang, and Geoffrey~J. Hay, editors, {\em Object-{{Based Image Analysis}}: {{Spatial Concepts}} for {{Knowledge-Driven Remote Sensing Applications}}}, pages 531--553. Springer, Berlin, Heidelberg, 2008.

\bibitem{kohliOntologySlumsImagebased2012}
Divyani Kohli, Richard Sliuzas, Norman Kerle, and Alfred Stein.
\newblock An ontology of slums for image-based classification.
\newblock {\em Computers, Environment and Urban Systems}, 36(2):154--163, March 2012.

\bibitem{kufferExtractionSlumAreas2016}
Monika Kuffer, Karin Pfeffer, Richard Sliuzas, and Isa Baud.
\newblock Extraction of {{Slum Areas From VHR Imagery Using GLCM Variance}}.
\newblock {\em IEEE Journal of Selected Topics in Applied Earth Observations and Remote Sensing}, 9(5):1830--1840, May 2016.

\bibitem{wurmExploitationTexturalMorphological2017}
Michael Wurm, Matthias Weigand, Andreas Schmitt, Christian Gei{\ss}, and Hannes Taubenb{\"o}ck.
\newblock Exploitation of textural and morphological image features in {{Sentinel-2A}} data for slum mapping.
\newblock In {\em 2017 {{Joint Urban Remote Sensing Event}} ({{JURSE}})}, pages 1--4, March 2017.

\bibitem{alzubaidiReviewDeepLearning2021}
Laith Alzubaidi, Jinglan Zhang, Amjad~J. Humaidi, Ayad {Al-Dujaili}, Ye~Duan, Omran {Al-Shamma}, J.~Santamar{\'i}a, Mohammed~A. Fadhel, Muthana {Al-Amidie}, and Laith Farhan.
\newblock Review of deep learning: Concepts, {{CNN}} architectures, challenges, applications, future directions.
\newblock {\em Journal of Big Data}, 8(1):53, March 2021.

\bibitem{perselloDeepFullyConvolutional2017}
Claudio Persello and Alfred Stein.
\newblock Deep {{Fully Convolutional Networks}} for the {{Detection}} of {{Informal Settlements}} in {{VHR Images}}.
\newblock {\em IEEE Geoscience and Remote Sensing Letters}, 14(12):2325--2329, December 2017.

\bibitem{hafnerUnsupervisedDomainAdaptation2022}
Sebastian Hafner, Yifang Ban, and Andrea Nascetti.
\newblock Unsupervised domain adaptation for global urban extraction using {{Sentinel-1 SAR}} and {{Sentinel-2 MSI}} data.
\newblock {\em Remote Sensing of Environment}, 280:113192, October 2022.

\bibitem{starkSlumMappingImbalanced2019}
Thomas Stark, Michael Wurm, Hannes Taubenb{\"o}ck, and Xiao~Xiang Zhu.
\newblock Slum {{Mapping}} in {{Imbalanced Remote Sensing Datasets Using Transfer Learned Deep Features}}.
\newblock In {\em 2019 {{Joint Urban Remote Sensing Event}} ({{JURSE}})}, pages 1--4, May 2019.

\bibitem{dengImageNetLargescaleHierarchical2009}
Jia Deng, Wei Dong, Richard Socher, Li-Jia Li, Kai Li, and Li~{Fei-Fei}.
\newblock {{ImageNet}}: {{A}} large-scale hierarchical image database.
\newblock In {\em 2009 {{IEEE Conference}} on {{Computer Vision}} and {{Pattern Recognition}}}, pages 248--255, June 2009.

\bibitem{fullerCROMARemoteSensing2023}
Anthony Fuller, Koreen Millard, and James~R. Green.
\newblock {{CROMA}}: {{Remote Sensing Representations}} with {{Contrastive Radar-Optical Masked Autoencoders}}.
\newblock https://arxiv.org/abs/2311.00566v1, November 2023.

\bibitem{rw2019timm}
Ross Wightman.
\newblock Pytorch image models.
\newblock https://github.com/rwightman/pytorch-image-models, 2019.

\bibitem{oquabDINOv2LearningRobust2023}
Maxime Oquab, Timoth{\'e}e Darcet, Th{\'e}o Moutakanni, Huy Vo, Marc Szafraniec, Vasil Khalidov, Pierre Fernandez, Daniel Haziza, Francisco Massa, Alaaeldin {El-Nouby}, Mahmoud Assran, Nicolas Ballas, Wojciech Galuba, Russell Howes, Po-Yao Huang, Shang-Wen Li, Ishan Misra, Michael Rabbat, Vasu Sharma, Gabriel Synnaeve, Hu~Xu, Herv{\'e} Jegou, Julien Mairal, Patrick Labatut, Armand Joulin, and Piotr Bojanowski.
\newblock {{DINOv2}}: {{Learning Robust Visual Features}} without {{Supervision}}.
\newblock https://arxiv.org/abs/2304.07193v2, April 2023.

\bibitem{wangAdvancingPlainVision2022}
Di~Wang, Qiming Zhang, Yufei Xu, Jing Zhang, Bo~Du, Dacheng Tao, and Liangpei Zhang.
\newblock Advancing {{Plain Vision Transformer Towards Remote Sensing Foundation Model}}.
\newblock https://arxiv.org/abs/2208.03987v4, August 2022.

\bibitem{chaBillionscaleFoundationModel2023}
Keumgang Cha, Junghoon Seo, and Taekyung Lee.
\newblock A {{Billion-scale Foundation Model}} for {{Remote Sensing Images}}, April 2023.

\bibitem{guoSkySenseMultiModalRemote2023}
Xin Guo, Jiangwei Lao, Bo~Dang, Yingying Zhang, Lei Yu, Lixiang Ru, Liheng Zhong, Ziyuan Huang, Kang Wu, Dingxiang Hu, Huimei He, Jian Wang, Jingdong Chen, Ming Yang, Yongjun Zhang, and Yansheng Li.
\newblock {{SkySense}}: {{A Multi-Modal Remote Sensing Foundation Model Towards Universal Interpretation}} for {{Earth Observation Imagery}}.
\newblock https://arxiv.org/abs/2312.10115v2, December 2023.

\bibitem{hongSpectralGPTSpectralRemote2023}
Danfeng Hong, Bing Zhang, Xuyang Li, Yuxuan Li, Chenyu Li, Jing Yao, Naoto Yokoya, Hao Li, Pedram Ghamisi, Xiuping Jia, Antonio Plaza, Paolo Gamba, Jon~Atli Benediktsson, and Jocelyn Chanussot.
\newblock {{SpectralGPT}}: {{Spectral Remote Sensing Foundation Model}}, November 2023.

\bibitem{heMaskedAutoencodersAre2021}
Kaiming He, Xinlei Chen, Saining Xie, Yanghao Li, Piotr Doll{\'a}r, and Ross Girshick.
\newblock Masked {{Autoencoders Are Scalable Vision Learners}}.
\newblock https://arxiv.org/abs/2111.06377v3, November 2021.

\bibitem{wangSSL4EOS12LargeScaleMultiModal2022}
Yi~Wang, Nassim Ait~Ali Braham, Zhitong Xiong, Chenying Liu, Conrad~M. Albrecht, and Xiao~Xiang Zhu.
\newblock {{SSL4EO-S12}}: {{A Large-Scale Multi-Modal}}, {{Multi-Temporal Dataset}} for {{Self-Supervised Learning}} in {{Earth Observation}}.
\newblock https://arxiv.org/abs/2211.07044v2, November 2022.

\bibitem{breimanRandomForests2001a}
Leo Breiman.
\newblock Random {{Forests}}.
\newblock {\em Machine Learning}, 45(1):5--32, October 2001.

\bibitem{fernandez-delgadoWeNeedHundreds2014a}
Manuel {Fern{\'a}ndez-Delgado}, Eva Cernadas, Sen{\'e}n Barro, and Dinani Amorim.
\newblock Do we need hundreds of classifiers to solve real world classification problems?
\newblock {\em J. Mach. Learn. Res.}, 15(1):3133--3181, January 2014.

\bibitem{heLearningImbalancedData2009}
Haibo He and Edwardo~A. Garcia.
\newblock Learning from {{Imbalanced Data}}.
\newblock {\em IEEE Transactions on Knowledge and Data Engineering}, 21(9):1263--1284, September 2009.

\end{thebibliography}

\end{document}